\documentclass{article} % For LaTeX2e

\usepackage{microtype}
\usepackage{hyperref}
\usepackage{url}
\usepackage{booktabs}
\usepackage[final]{colm2025_xllmworkshop}
\usepackage{lineno}

\usepackage{xcolor}         % colors
\usepackage{amsmath}
\usepackage{cleveref}
\usepackage{graphicx} 
\usepackage{CJKutf8}
\usepackage{subcaption}
\usepackage{float}
\usepackage{adjustbox}
\usepackage{multirow}
\definecolor{darkblue}{rgb}{0, 0, 0.5}
\hypersetup{colorlinks=true, citecolor=darkblue, linkcolor=darkblue, urlcolor=darkblue}
\usepackage{tcolorbox}
\usepackage{listings}
\tcbuselibrary{listings}
\usepackage[T1]{fontenc}
\usepackage[utf8]{inputenc}
\usepackage{newtxtt}  % Optional: cleaner monospace font
\usepackage{tabularx}

\title{Latent Chain-of-Thought? Decoding the Depth-Recurrent Transformer}

\author{Wenquan Lu$^{1}$, Yuechuan Yang$^{1}$, Kyle Lee$^{1}$, Yanshu Li$^{1}$, Enqi Liu$^{2}$\\
\textsuperscript{1}Brown University, \textsuperscript{2}Harvard University\\
\texttt{wenquan\_lu@brown.edu} \\
}

\begin{document}

\ifcolmsubmission
\linenumbers
\fi

\maketitle

\begin{abstract}
Chain-of-thought (CoT) reasoning has enabled transformer-based language models to excel at complex mathematics and multi-step planning. However, in standard decoder-only architectures, these reasoning steps are externalized in natural language, improving interpretability at the cost of efficiency. To capture reasoning that is not easily represented in words, many works have explored recurrent architectures that aim to internalize reasoning in latent space, potentially supporting latent CoT. In this paper, we investigate whether such reasoning structures emerge in Huginn-3.5B, a depth-recurrent Transformer that reuses layers at inference time without increasing parameter count. We examine the model's internal behavior on arithmetic tasks using a suite of probing techniques including the Logit Lens and Coda Lens. Our findings reveal limited evidence of interpretable latent CoT by tracking rank trajectories of final and intermediate result tokens. Furthermore, we uncover significant probing inconsistencies across recurrent blocks, where the interpretability of hidden states depends heavily on both the layer index and the decoding method. Finally, we empirically show that increasing recurrence depth yields only marginal gains and falls well short of models that explicitly externalize reasoning steps. The code is available at \url{https://github.com/wenquanlu/huginn-latent-cot}.
\end{abstract}

\section{Introduction}
Modern large language models demonstrate remarkable capabilities in reasoning and planning tasks~\citep{guo2025deepseek, rela7}. Much of this success relies on Chain-of-thought \citep{wei2022chain, rela1, rela2}: explicitly prompting the model to articulate intermediate steps in natural language. This strategy, though effective, may introduce verbosity and slow inference. A compelling alternative is to develop models that perform reasoning in latent space without surfacing intermediate steps in language. Yet, it remains unclear whether today's architectures are capable of such behavior.

A promising approach to latent reasoning leverages recurrent methods~\citep{coconut}, where intermediate continuous hidden states are passed across reused layers to simulate multi-step reasoning without emitting language. The Huginn-3.5B model~\citep{geiping2025scaling} exemplifies this idea with a depth-recurrent Transformer that reuses layers at inference to increase computational depth per token. While increasing recurrences improves performance on reasoning tasks, it remains unclear whether this stems from iterative refinement or the emergence of structured, CoT-like reasoning in latent space~\citep{rela21}.

As a result, in this paper, we ask: \textit{Does Huginn exhibit signs of latent chain-of-thought reasoning during inference?} To investigate this, we conduct a systematic analysis of Huginn's hidden states on arithmetic tasks under conditions that suppress explicit reasoning. We introduce an unrolled view of the architecture and apply a range of probing techniques including logit lens, coda lens, and token rank trajectory tracking to decode and visualize the model's internal computations. In summary, our key contributions and findings are as follows.

1. \textbf{We uncover significant probing inconsistencies across blocks in Huginn's depth-recurrent architecture.} Unlike the smoothly evolving representations typically observed in feedforward Transformers~\citep{nostalgebraist2021logitlens}, Huginn exhibits sharp discontinuities in hidden state semantics across layers. In particular, different blocks (e.g., R1 vs. R4) encode distinct information, and their interpretability strongly depends on the choice of decoding lens.
2. \textbf{Token rank trajectory analysis provides little evidence for latent CoT reasoning.} By tracing the rank dynamics of both intermediate and final answer tokens, we find no clear temporal separation or structured latent reasoning pathway across recurrence steps, contrary to what latent CoT would predict.
3. \textbf{Scaling recurrence depth fails to match explicit reasoning.} On the GSM8K benchmark, increasing the number of recurrent steps only marginally improves performance and falls far short of models that leverage explicit CoT prompting, highlighting the limitations of depth recurrence \textit{alone} for inducing effective reasoning behavior.

\begin{figure*}[t]
    \centering
    \includegraphics[width =0.8\columnwidth]{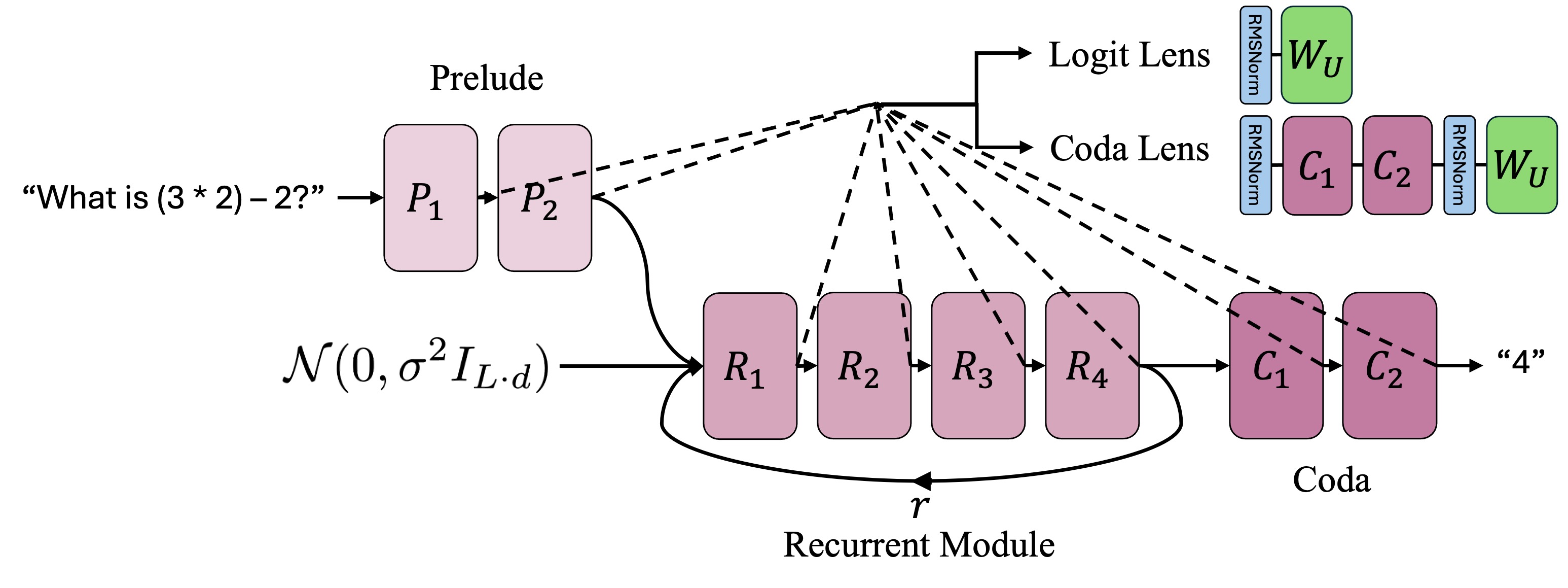}
    \vspace{-10pt}
    \caption{Overview of our approach to decode hidden states in the depth-recurrent Huginn. For each block's output, we employ logit lens and coda lens to convert them into logits.}
    \vspace{-5pt}
    \label{fig:arch}
\end{figure*}

\section{Method}
In this section, we first introduce an unrolled view of the Huginn architecture. We then present the two main approaches for probing and decoding the hidden states of the depth-recurrent transformer to detect latent chain-of-thought (CoT): logit lens and coda lens.

\subsection{Unrolled View of Huginn Architecture}
As shown in \cref{fig:arch}, the architecture of Huginn 3.5B model consists of 2 Prelude blocks $\{P_1, P_2\}$, 4 Recurrent Blocks $\{R_1, R_2, R_3, R_4\}$ and 2 Coda blocks $\{C_1, C_2\}$, where each block is a standard, causal self-attention block. Given input tokens $x\in \mathbb{R}^{L\times |V|}$, the input is first embedded by the embedding matrix $W_E\in \mathbb{R}^{|V|\times d}$ to input embeddings $e = xW_{E}, e\in \mathbb{R}^{L\times d}$. These embeddings are first processed by the Prelude blocks, followed by $r$ recurrent passes through the Recurrent blocks, and finally the Coda blocks for prediction. By unrolling the recurrence, the input embeddings are passed through $2 + 4r + 2$ blocks, where $r$ denotes the number of recurrent steps taken during a single forward pass for next-token prediction. The hidden states $s_i$ produced by each block can be summarized in the following equation:
\begin{align}
s_i = 
\begin{cases}
  e & i = 0 \\
  P_i(s_{i-1}) & 1 \le i \le 2 \\
  R_1(s_2, n),\quad n \sim \mathcal{N}(0, \sigma^2 I_{L\cdot d}) & i = 3 \\
  R_{(i - 3) \bmod 4 + 1}(s_{i-1}) & 4 \le i \le 2 + 4r,\ i \not\equiv 3 \pmod{4} \\
  R_1(s_2, s_{i-1}) & 4 \le i \le 2 + 4r,\ i \equiv 3 \pmod{4} \\
  C_{i - (2 + 4r)}(s_{i-1}) & 2 + 4r + 1 \le i \le 2 + 4r + 2
\end{cases}
\label{eq:unroll}
\end{align}
Note that in the third line of \cref{eq:unroll}, a random vector drawn from normal distribution is used as the initial state for the recurrence. From this unrolled perspective, there are $2 + 4r + 2$ hidden states which we can track the trajectory of intermediate computations. 
\subsection{Decoding Hidden States by Logit Lens and Coda Lens}
\textbf{Logit lens.} The logit lens is a widely used technique for interpreting intermediate representations in transformer-based models. As illustrated in~\cref{fig:arch}, for each hidden state $s_i$, we first apply RMS normalization, followed by projection through the unembedding matrix $W_U \in \mathbb{R}^{d \times |V|}$ to obtain logits $z_i$ in the vocabulary space. These steps align with Huginn's architectural choice of normalizing features prior to unembedding:
\begin{align}
    z_i = \mathrm{RMSNorm}(s_i) W_U
\end{align}
 We focus on the logits from the last token position $z_i[-1]$ which associates with model's current prediction. Prior work has shown that such projections often yield interpretable, top-ranked tokens that are aligned with the model's internal computation~\citep{geva-etal-2023-dissecting, geva-etal-2022-transformer}. Thus, analyzing $z_i[-1]$ across intermediate layers or recurrent steps provides insight into how the model's predictions evolve over time.

\textbf{Coda Lens}. The Huginn model's recurrent architecture includes a specialized coda module $C = \{C_1, C_2\}$ to effectively map the output of the last recurrent block to logits during inference. The coda block is a more expressive decoder compared to logit lens as it consists of two transformer blocks. Accordingly, we also decode the hidden states using this learned module, a method we refer to as the \textit{coda lens}, to explore whether it produces more faithful or semantically aligned logits:
\begin{align}
    z_i = \mathrm{RMSNorm}(C(\mathrm{RMSNorm}(s_i))) W_U
\end{align}
In line with Huginn's implementation, we apply normalization both before and after the coda module to ensure consistency. As with the logit lens, we focus our analysis on $z_i[-1]$.

\section{Experiment}

\subsection{Experimental Setup for Mathematical Reasoning Without Explicit CoT}
\textbf{Datasets.} 
In our main experiment, we use the synthetic arithmetic test data employed in GPT-3 \citep{gpt3}. Specifically, We use the \textbf{one-digit composite} task: the model is asked to perform a composite operation on three 1 digit numbers. For example, "Question: What is (9 + 8) * 2? Answer: 34". We use such a simple dataset for evaluation and analysis because the Huginn model only achieves an accuracy of 0.19 on the dataset. The dataset has in total 2k questions. To ensure negative results are not artifacts of errors, we later restrict our analysis to correctly answered subsets. In addition, we also test model's performance on the standard mathematical reasoning dataset GSM8K~\citep{cobbe2021training} which contains 8.5K high quality grade school math word problems.

\textbf{Suppress Explicit CoT.} To encourage latent chain-of-thought (CoT), in all experiments, we suppress explicit CoT by enforcing the model to output the answer straightway using system message and four in-context examples (see \cref{app:suppress}). We format system message as "You are a concise and helpful assistant. Always return only the final answer straightway." In in-context examples, the final answer of the question are given as correct output without any additional token. Unless otherwise mentioned, we set recurrent steps to 16 for all experiments. So in total there are $2 + 4 \times 16 + 2 = 68$ blocks.

\subsection{Discontinuities in Hidden State Interpretability of Depth-Recurrent Transformer}\label{sec:decode}
\begin{figure}[ht]
    \centering
    \begin{minipage}[t]{0.49\textwidth}
        \centering
        \includegraphics[width=\linewidth]{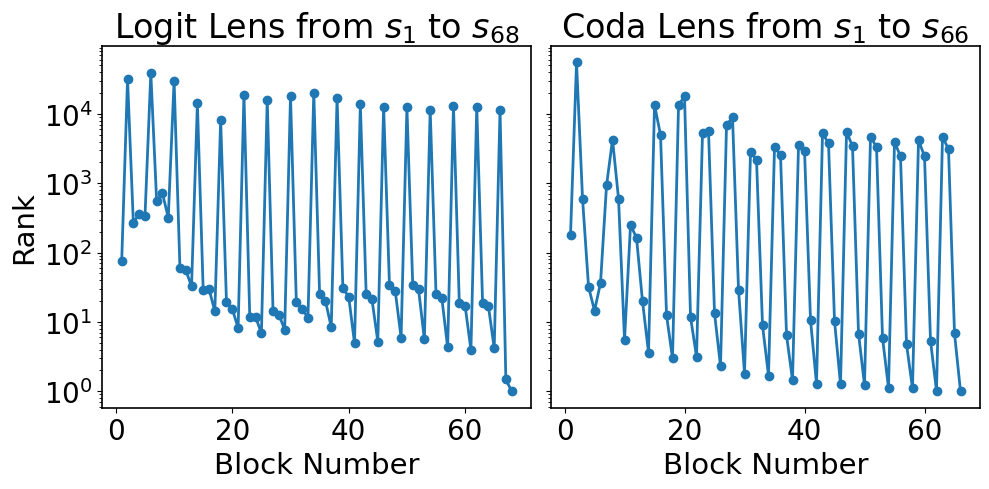}
        \caption{Average rank trajectory of the final predicted token (via logit lens (left), coda lens (right)) across unrolled blocks, averaged over 100 arithmetic questions. Note the first two ranks in each graph are from the prelude.}
        \label{fig:main_unrolled}
    \end{minipage}
    \hfill
    \begin{minipage}[t]{0.49\textwidth}
        \centering
        \includegraphics[width=\linewidth]{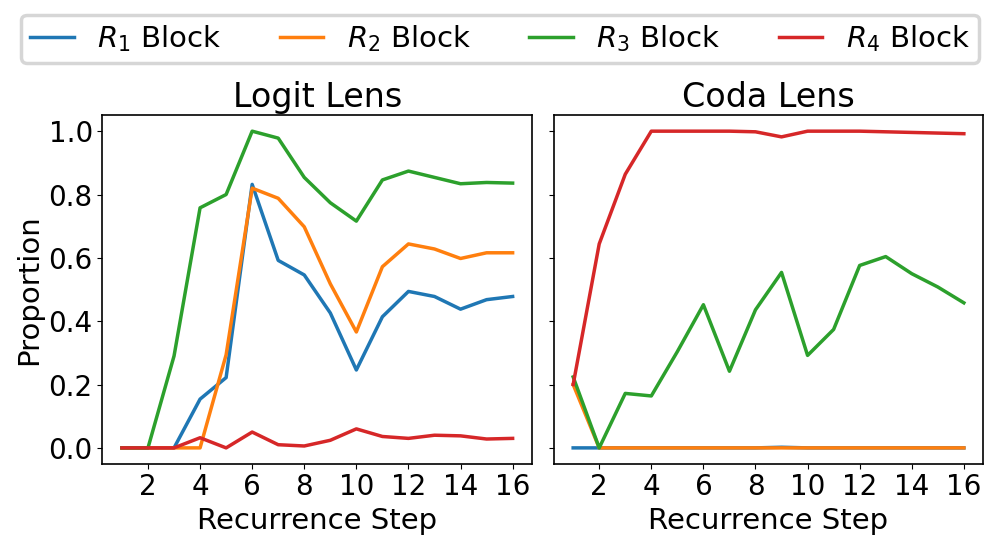}
        \caption{Proportion of top-5 decoded tokens (via logit lens (left), coda lens (right)) that are valid signed-integer prefixes across recurrence steps, averaged over 100 arithmetic questions.}
        \label{fig:top5_unrolled}
    \end{minipage}
\end{figure}

As a first step in our investigation, we examine whether the outputs decoded by the logit lens and coda lens across Huginn's unrolled layers exhibit the pattern of \textbf{initial guess followed by smooth refinement} observed in conventional decoder-only language models~\citep{vaswani2017attention}, a phenomenon originally identified through logit lens analysis, where models quickly reach a coarse prediction in early layers, then progressively converge toward the final prediction~\citep{nostalgebraist2021logitlens}. Since Huginn is a depth-recurrent transformer, it is unclear whether such interpretability patterns still hold. If this pattern were observed, it would provide strong evidence against the presence of latent CoT dynamics, as smooth refinement implies continuous convergence rather than discrete internal transitions. However, as we show below, Huginn exhibits markedly different behavior. To test this, we select the first 100 arithmetic questions from the arithmetic dataset and run Huginn forward \textbf{once} on each question, recording the average rank of the final predicted tokens decoded from each unrolled block. The trajectories of ranks through unrolled layers are visualized in \cref{fig:main_unrolled}.

As seen from \cref{fig:main_unrolled}, the ranks exhibit large-magnitude, periodic oscillations as the token moves through layers. Note that we use logarithmic scale for vertical axis as the rank distributions vary substantially across layers. In \cref{fig:main_unrolled} (left), we observe consistent upward spikes at the $R_4$ layer, while most other layers exhibit low decoded ranks. This pattern reverses in \cref{fig:main_unrolled} (right), where the downward spikes consistently occur at $R_4$ layer, with high decoded ranks concentrated in the other layers.

The observation is further supported by \cref{fig:top5_unrolled}, where we measure the proportion of top-5 decoded tokens that are valid signed-integer prefixes (e.g., "5", "-", "2"). Because the 100 test questions are purely arithmetic, these prefixes are expected intermediate steps if the model engages in latent CoT reasoning. As shown in \cref{fig:top5_unrolled} (left), a large proportion of top decoded tokens from blocks $\{R_1, R_2, R_3\}$ using the logit lens are signed numeric prefixes, while almost none from $R_4$ are. In contrast, for the coda lens \cref{fig:top5_unrolled} (right), nearly 100\% of tokens decoded from $R_4$ are signed numeric prefixes, while $\{R_1, R_2\}$ yield virtually none.

Manual inspection of top-5 tokens further illustrates the divergence both across blocks and between decoding methods. For example, $R_4$ decoded with logit lens produces uninterpretable outputs such as \{\texttt{"inc"}, \texttt{"unity"}, \texttt{"friendships"}, \texttt{"igne"}, \texttt{"impulse"}\}, whereas coda lens on exactly the same hidden state produces numerical tokens that closely relates to the arithmetic computation: \{\texttt{"6"}, \texttt{"5"}, \texttt{"1"}, \texttt{"7"}, \texttt{"2"}\}. \textbf{Conversely}, at the earlier block $R_1$, decoding with coda lens produces general-purpose outputs such as \{\texttt{"answer"}, \texttt{"answers"}, \texttt{"tru"}, \texttt{"clarification"}, \texttt{"spa"}\}. These tokens relate to the semantics of answering in general, rather than reflecting numerical computations. However, decoding via logit lens from exactly the same hidden state still results in a high proportion of numerical tokens: \{\texttt{"5"}, \texttt{"3"}, \texttt{"1"}, \texttt{" answer"}, \texttt{"None"}\}. Further examples are provided in \cref{app:examples}.

These discoveries suggest that the interpretability of intermediate states in Huginn, when probed using logit and coda lens, varies dramatically depending on the block index and decoding method. There are clear discontinuities in representations across blocks, particularly around $R_4$. This shows top decoded tokens do not form a smooth convergence toward the final prediction over unrolled layers, and that \textbf{lens applicability must be assessed on a per-layer basis}. One possible explanation for the distinct behavior of $R_4$ is that its output can serve a dual role: feeding into both the next recurrent cycle via $R_1$ and the coda $C_1$, which may force it to encode a representation that differs markedly from other blocks.

\subsection{Tracing Final and Intermediate Tokens Provides Little Evidence for Latent CoT}\label{sec:trace}
We now investigate whether Huginn exhibits latent CoT by tracing the rank trajectories of \textbf{signature tokens}: the intermediate and final result tokens in arithmetic problems. These tokens serve as anchors for assessing whether the model performs multi-step reasoning across recurrence. In our setting of 1-digit composite task, e.g., 2 * 3 + 1, the intermediate result is 6, and the final result is 7. We filter the 2k arithmetic dataset to a subset of 67 questions using the following criteria (1) the model predicts correct answer. (2) the final result and any intermediate result are only single digit/token; the constraint improves the interpretability because the language model operates on token-level information. (3) the final result is different to intermediate result so their rank trajectory do not trivially overlap. 

To address the blockwise inconsistency discovered in \cref{sec:decode} and prevent drastic oscillatons in graph, we restrict our analysis and visualization to $R_3$ outputs for logit lens, and $R_4$ outputs for coda lens. These blocks have best alignment with final prediction and can be decoded into interpretable domain as shown in \cref{sec:decode}.

\cref{fig:inter_block} shows the rank trajectory of the final result token and intermediate result token across recurrent steps, we also include the rank of a random token `the' as a baseline reference. If the model performs latent CoT, we expect the rank of the intermediate token to drop first, followed by a delayed drop in the final token's rank, reflecting stepwise reasoning. However, such phase separation is not observed in \cref{fig:inter_block}. In both subplots, the ranks of both the final and intermediate tokens descend quickly in early recurrent steps, with the final token consistently maintaining a lower rank than the intermediate token. Hence, no clear evidence of latent CoT is observed based on the rank analysis. However, an interesting \textbf{rank reversal} between final and intermediate tokens at around step 6 is observed for most examples. This could potentially indicate the model is re-evaluating the final outcome based on intermediate results. We leave a deeper investigation of this phenomenon to future work. The graphs for other recurrent blocks are provided in \cref{app:further_r}, which also do not signal any latent CoT. 
\vspace{-5pt}
\begin{minipage}[c]{0.48\textwidth}
\vspace{-5pt}
    \begin{figure}[H]  % or [ht] if you want LaTeX to float it
        \centering
        \includegraphics[width=\linewidth]{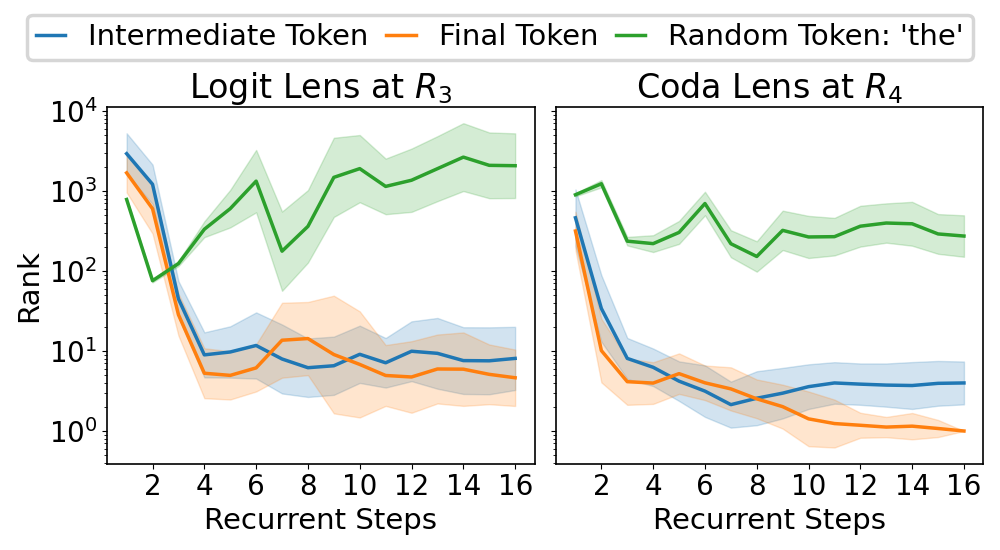}
        \vspace{-15pt}
        \caption{Rank trajectory of the final, intermediate, and random token (decoded via logit lens (left), coda lens (right)), averaged over 67 single-digit arithmetic questions that the model answers correctly. Shaded regions denote $\pm 1$ relative std.}
        \label{fig:inter_block}
    \end{figure}
\end{minipage}
\hfill
\begin{minipage}[c]{0.48\textwidth}
\vspace{-5pt}
    \begin{table}[H]  % or [ht]
        \centering
        \adjustbox{max width=\linewidth}{
        \begin{tabular}{l|c|c}
            \hline
            \textbf{Model} & \textbf{Recurrent Steps} & \textbf{GSM8K Accuracy} \\
            \hline
            Huginn & 64 & \textbf{24.87}/\textbf{38.13} \\
            \hline
            \multirow{7}{*}{Huginn w/o CoT}
            & 4 & 3.11/3.11 \\
            & 8 & 4.47/4.47 \\
            & 16 & 4.78/4.78 \\
            & 32 & 4.93/4.93 \\
            & 64 & 4.70/4.70 \\
            & 128 & 4.93/4.93 \\
            & 256 & 4.62/4.62 \\
            \hline
        \end{tabular}
        }
        \caption{GSM8K accuracy (strict/flexible) across different models and recurrence steps. Without explicit CoT, there is a monotonic increase in accuracy as recurrent steps increases from 4 to 32. However, it is still substantially lower than that with explicit CoT as shown in the first row of the table.}
        \label{tab:gsm8k_accuracy}
    \end{table}
\end{minipage}

\subsection{Scaling Recurrent Steps Cannot Beat Explicit Chain-of-Thought}\label{sec:gsm8k}
Given the lack of clear evidence for latent chain-of-thought (CoT) reasoning in our earlier probing analysis, we turn to a macroscopic performance evaluation to detect any indirect traces of such behavior. Specifically, we benchmark Huginn on the GSM8K dataset. As in prior experiments, we suppress explicit CoT reasoning by modifying the system message. To remain consistent with the original Huginn evaluation setup, we use an 8-shot prompting format. As shown in \cref{tab:gsm8k_accuracy}, increasing the number of recurrent steps from 4 to 32 leads to only modest gains in accuracy (from 3.11 to 4.93), and performance plateaus thereafter. In contrast, Huginn with explicit CoT achieves significantly higher accuracy (24.87/38.13). This suggests that even if some latent reasoning emerges within the recurrent loop, it is insufficient to rival the effectiveness of standard chain-of-thought reasoning.
\vspace{-5pt}
\section{Conclusion}
\vspace{-5pt}
In this paper, we investigated whether depth-recurrent transformers, exemplified by Huginn, exhibit latent chain-of-thought reasoning. We analyze the internal dynamics of the model on arithmetic tasks under conditions that suppress explicit reasoning. Through logit lens and coda lens, we find little evidence of structured latent chain-of-thought reasoning. However, our results do not definitively rule out the presence of latent CoT. If it exists, it may be more subtle or distributed than our employed tools can detect. Future work may apply more advanced probing techniques, such as activation patching~\citep{meng2022locating}, to uncover finer-grained reasoning patterns potentially hidden within the recurrent loop. 

\bibliography{colm2025_conference}

\begin{thebibliography}{15}
\providecommand{\natexlab}[1]{#1}
\providecommand{\url}[1]{\texttt{#1}}
\expandafter\ifx\csname urlstyle\endcsname\relax
  \providecommand{\doi}[1]{doi: #1}\else
  \providecommand{\doi}{doi: \begingroup \urlstyle{rm}\Url}\fi

\bibitem[Brown et~al.(2020)Brown, Mann, Ryder, Subbiah, Kaplan, Dhariwal, Neelakantan, Shyam, Sastry, Askell, Agarwal, Herbert-Voss, Krueger, Henighan, Child, Ramesh, Ziegler, Wu, Winter, Hesse, Chen, Sigler, Litwin, Gray, Chess, Clark, Berner, McCandlish, Radford, Sutskever, and Amodei]{gpt3}
Tom~B. Brown, Benjamin Mann, Nick Ryder, Melanie Subbiah, Jared Kaplan, Prafulla Dhariwal, Arvind Neelakantan, Pranav Shyam, Girish Sastry, Amanda Askell, Sandhini Agarwal, Ariel Herbert-Voss, Gretchen Krueger, Tom Henighan, Rewon Child, Aditya Ramesh, Daniel~M. Ziegler, Jeffrey Wu, Clemens Winter, Christopher Hesse, Mark Chen, Eric Sigler, Mateusz Litwin, Scott Gray, Benjamin Chess, Jack Clark, Christopher Berner, Sam McCandlish, Alec Radford, Ilya Sutskever, and Dario Amodei.
\newblock Language models are few-shot learners.
\newblock 2020.

\bibitem[Chen et~al.(2024)Chen, Qin, Wang, Zhou, and Che]{rela2}
Qiguang Chen, Libo Qin, Jiaqi Wang, Jingxuan Zhou, and Wanxiang Che.
\newblock Unlocking the capabilities of thought: A reasoning boundary framework to quantify and optimize chain-of-thought.
\newblock \emph{Advances in Neural Information Processing Systems}, 37:\penalty0 54872--54904, 2024.

\bibitem[Cobbe et~al.(2021)Cobbe, Kosaraju, Bavarian, Chen, Jun, Kaiser, Plappert, Tworek, Hilton, Nakano, et~al.]{cobbe2021training}
Karl Cobbe, Vineet Kosaraju, Mohammad Bavarian, Mark Chen, Heewoo Jun, Lukasz Kaiser, Matthias Plappert, Jerry Tworek, Jacob Hilton, Reiichiro Nakano, et~al.
\newblock Training verifiers to solve math word problems.
\newblock \emph{arXiv preprint arXiv:2110.14168}, 2021.

\bibitem[Geiping et~al.(2025)Geiping, McLeish, Jain, Kirchenbauer, Singh, Bartoldson, Kailkhura, Bhatele, and Goldstein]{geiping2025scaling}
Jonas Geiping, Sean McLeish, Neel Jain, John Kirchenbauer, Siddharth Singh, Brian~R Bartoldson, Bhavya Kailkhura, Abhinav Bhatele, and Tom Goldstein.
\newblock Scaling up test-time compute with latent reasoning: A recurrent depth approach.
\newblock \emph{arXiv preprint arXiv:2502.05171}, 2025.

\bibitem[Geva et~al.(2022)Geva, Caciularu, Wang, and Goldberg]{geva-etal-2022-transformer}
Mor Geva, Avi Caciularu, Kevin Wang, and Yoav Goldberg.
\newblock Transformer feed-forward layers build predictions by promoting concepts in the vocabulary space.
\newblock In Yoav Goldberg, Zornitsa Kozareva, and Yue Zhang (eds.), \emph{Proceedings of the 2022 Conference on Empirical Methods in Natural Language Processing}, pp.\  30--45, Abu Dhabi, United Arab Emirates, December 2022. Association for Computational Linguistics.
\newblock \doi{10.18653/v1/2022.emnlp-main.3}.
\newblock URL \url{https://aclanthology.org/2022.emnlp-main.3/}.

\bibitem[Geva et~al.(2023)Geva, Bastings, Filippova, and Globerson]{geva-etal-2023-dissecting}
Mor Geva, Jasmijn Bastings, Katja Filippova, and Amir Globerson.
\newblock Dissecting recall of factual associations in auto-regressive language models.
\newblock In Houda Bouamor, Juan Pino, and Kalika Bali (eds.), \emph{Proceedings of the 2023 Conference on Empirical Methods in Natural Language Processing}, pp.\  12216--12235, Singapore, December 2023. Association for Computational Linguistics.
\newblock \doi{10.18653/v1/2023.emnlp-main.751}.
\newblock URL \url{https://aclanthology.org/2023.emnlp-main.751/}.

\bibitem[Guo et~al.(2025)Guo, Yang, Zhang, Song, Zhang, Xu, Zhu, Ma, Wang, Bi, et~al.]{guo2025deepseek}
Daya Guo, Dejian Yang, Haowei Zhang, Junxiao Song, Ruoyu Zhang, Runxin Xu, Qihao Zhu, Shirong Ma, Peiyi Wang, Xiao Bi, et~al.
\newblock Deepseek-r1: Incentivizing reasoning capability in llms via reinforcement learning.
\newblock \emph{arXiv preprint arXiv:2501.12948}, 2025.

\bibitem[Hao et~al.(2024)Hao, Sukhbaatar, Su, Li, Hu, Weston, and Tian]{coconut}
Shibo Hao, Sainbayar Sukhbaatar, DiJia Su, Xian Li, Zhiting Hu, Jason Weston, and Yuandong Tian.
\newblock Training large language models to reason in a continuous latent space.
\newblock \emph{arXiv preprint arXiv:2412.06769}, 2024.

\bibitem[Meng et~al.(2022)Meng, Bau, Andonian, and Belinkov]{meng2022locating}
Kevin Meng, David Bau, Alex Andonian, and Yonatan Belinkov.
\newblock Locating and editing factual associations in gpt.
\newblock \emph{Advances in neural information processing systems}, 35:\penalty0 17359--17372, 2022.

\bibitem[nostalgebraist(2021)]{nostalgebraist2021logitlens}
nostalgebraist.
\newblock Interpreting gpt: the logit lens, 2021.
\newblock URL \url{https://www.lesswrong.com/posts/AcKRB8wDpdaN6v6ru/interpreting-gpt-the-logit-lens}.
\newblock Accessed: 2025-03-21.

\bibitem[Vaswani et~al.(2017)Vaswani, Shazeer, Parmar, Uszkoreit, Jones, Gomez, Kaiser, and Polosukhin]{vaswani2017attention}
Ashish Vaswani, Noam Shazeer, Niki Parmar, Jakob Uszkoreit, Llion Jones, Aidan~N Gomez, {\L}ukasz Kaiser, and Illia Polosukhin.
\newblock Attention is all you need.
\newblock \emph{Advances in neural information processing systems}, 30, 2017.

\bibitem[Wei et~al.(2022)Wei, Wang, Schuurmans, Bosma, Xia, Chi, Le, Zhou, et~al.]{wei2022chain}
Jason Wei, Xuezhi Wang, Dale Schuurmans, Maarten Bosma, Fei Xia, Ed~Chi, Quoc~V Le, Denny Zhou, et~al.
\newblock Chain-of-thought prompting elicits reasoning in large language models.
\newblock \emph{Advances in neural information processing systems}, 35:\penalty0 24824--24837, 2022.

\bibitem[Yang et~al.(2024)Yang, Gribovskaya, Kassner, Geva, and Riedel]{rela21}
Sohee Yang, Elena Gribovskaya, Nora Kassner, Mor Geva, and Sebastian Riedel.
\newblock Do large language models latently perform multi-hop reasoning?
\newblock \emph{arXiv preprint arXiv:2402.16837}, 2024.

\bibitem[Yu et~al.(2023)Yu, Jiang, Shi, Yu, Liu, Zhang, Kwok, Li, Weller, and Liu]{rela7}
Longhui Yu, Weisen Jiang, Han Shi, Jincheng Yu, Zhengying Liu, Yu~Zhang, James~T Kwok, Zhenguo Li, Adrian Weller, and Weiyang Liu.
\newblock Metamath: Bootstrap your own mathematical questions for large language models.
\newblock \emph{arXiv preprint arXiv:2309.12284}, 2023.

\bibitem[Zhang et~al.(2024)Zhang, Du, Pang, Liu, Gao, and Lin]{rela1}
Xuan Zhang, Chao Du, Tianyu Pang, Qian Liu, Wei Gao, and Min Lin.
\newblock Chain of preference optimization: Improving chain-of-thought reasoning in llms.
\newblock \emph{Advances in Neural Information Processing Systems}, 37:\penalty0 333--356, 2024.

\end{thebibliography}
\bibliographystyle{colm2025_conference}

\appendix
\section{Appendix}
\subsection{Rank Trajectories of the Intermediate, Final and Random Tokens from Other Recurrent Blocks}\label{app:further_r}
\begin{figure}[H]
    \centering
    \includegraphics[width=\linewidth]{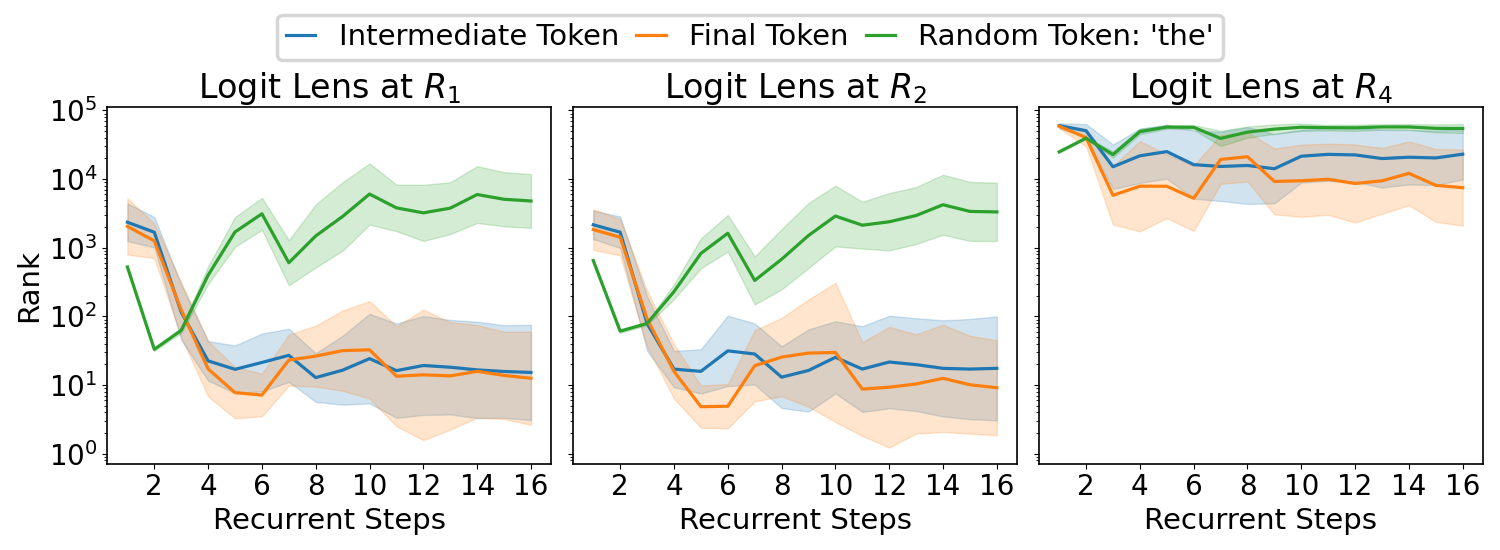}
    \caption{Rank trajectories of the intermediate, final and random tokens decoded by logit lens at recurrent blocks 1 (left), 2 (middle) and 4 (right). Shaded regions denote $\pm 1$ relative std.}
    \label{fig:enter-label}
\end{figure}
\begin{figure}[H]
    \centering
    \includegraphics[width=\linewidth]{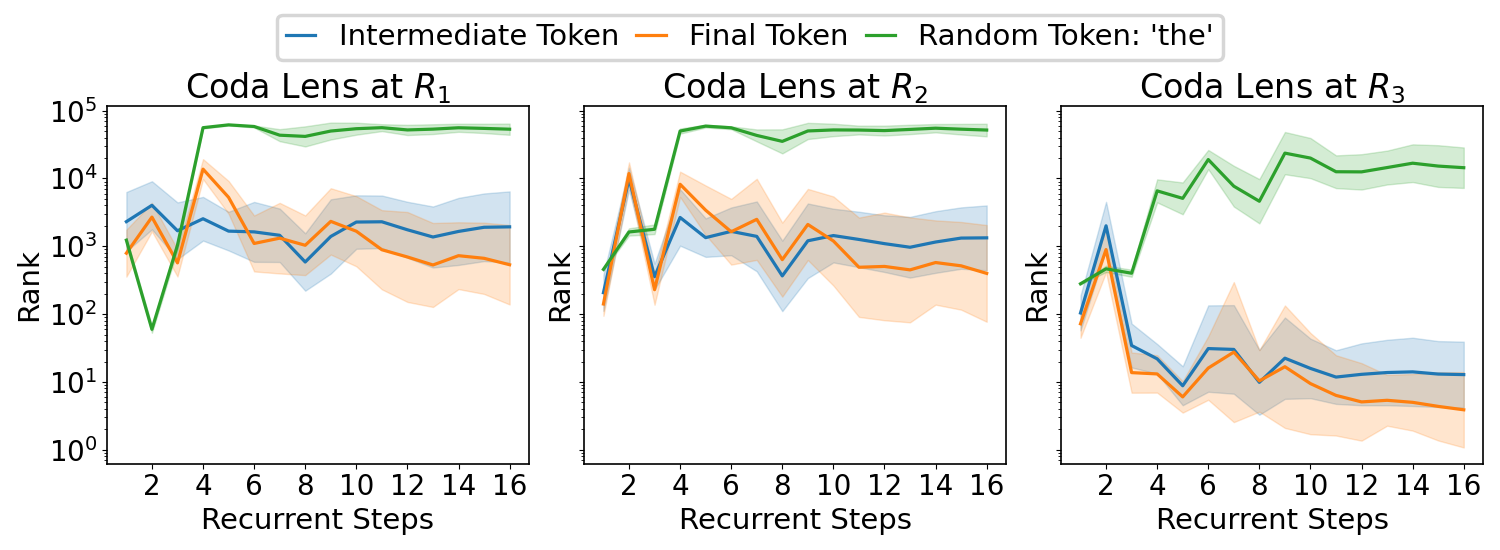}
    \caption{Rank trajectories of the Intermediate, final and random tokens decoded by coda lense at recurrent blocks 1 (left), 2(middle) and 3 (right). Shaded regions denote $\pm 1$ relative std.}
    \label{fig:enter-label}
\end{figure}

\subsection{Prompting with Suppressed CoT}\label{app:suppress}

\tcbset{
  colback=gray!5!white,
  colframe=gray!75!black,
  fonttitle=\bfseries,
  boxsep=3pt,
  arc=2pt,
  outer arc=2pt,
  left=5pt,
  right=5pt,
  top=5pt,
  bottom=5pt,
}
\begin{tcolorbox}[title=4-shot Prompt for \cref{sec:decode}, listing only, listing options={
    basicstyle=\ttfamily\scriptsize,
    breaklines=true,
    showstringspaces=false,
    escapeinside=||,
    literate={<}{<}1 {>}{>}1 {|}{|}1 {_}{\_}1,
}]
You are a concise and helpful assistant. Always return only the final answer straightway.

Question: What is (9 + 8) * 2?

Answer: 34

Question: What is (4 - 7) - 3?

Answer: -6

Question: What is (1 - 5) - 6?

Answer: -10

Question: What is (1 - 9) * 5?

Answer: -40
\end{tcolorbox}
%\vspace{-10pt}
\begin{tcolorbox}[title=4-shot Prompt for \cref{sec:trace}, listing only, listing options={
    basicstyle=\ttfamily\scriptsize,
    breaklines=true,
    showstringspaces=false,
    escapeinside=||,
    literate={<}{<}1 {>}{>}1 {|}{|}1 {_}{\_}1,
}]
You are a concise and helpful assistant. Always return only the final answer straightway.

Question: What is (5 + 1) + 1?

Answer: 7

Question: What is (2 + 5) - 1?

Answer: 6

Question: What is (6 - 4) + 5?

Answer: 7

Question: What is (2 + 4) - 1?

Answer: 5
\end{tcolorbox}
%\vspace{-10pt}

\begin{tcolorbox}[title=8-shot Prompt for \cref{sec:gsm8k}, listing only, listing options={
    basicstyle=\ttfamily\scriptsize,
    breaklines=true,
    showstringspaces=false,
    escapeinside=||,
    literate={<}{<}1 {>}{>}1 {|}{|}1 {_}{\_}1,
}]
You are a concise and helpful assistant. Always return only the final answer straightway.

There are 15 trees in the grove. Grove workers will plant trees in the grove today. After they are done, there will be 21 trees. How many trees did the grove workers plant today?

6

If there are 3 cars in the parking lot and 2 more cars arrive, how many cars are in the parking lot?

5

Leah had 32 chocolates and her sister had 42. If they ate 35, how many pieces do they have left in total?

39

Jason had 20 lollipops. He gave Denny some lollipops. Now Jason has 12 lollipops. How many lollipops did Jason give to Denny? 

8

Shawn has five toys. For Christmas, he got two toys each from his mom and dad. How many toys does he have now?

9

There were nine computers in the server room. Five more computers were installed each day, from monday to thursday. How many computers are now in the server room?

29

Michael had 58 golf balls. On tuesday, he lost 23 golf balls. On wednesday, he lost 2 more. How many golf balls did he have at the end of wednesday?

33

Olivia has \$23. She bought five bagels for \$3 each. How much money does she have left?

8
\end{tcolorbox}
For experiments in \cref{sec:decode}, \cref{sec:trace} and \cref{sec:gsm8k}, we use the above prompts to guide the model to generate the answer without explicit CoT reasoning. For \cref{sec:trace}, we additionally constrain the in-context examples to have single-digit answers, consistent with its experimental setup.

\subsection{Further Examples of the Decoded Top-5 Tokens}\label{app:examples}

Logit lens:
\begin{table}[H]  % or [ht]
        \centering
        \begin{tabularx}{\columnwidth}{|c|X|}
            \hline
            \textbf{Recurrent Block} & \textbf{Top-5 Logit Lens Decoded Tokens Examples} \\
            \hline
            \multirow{3}{*}{$R_1$} & \texttt{\char123`\ \ \ \ \ \ \ \ ' , `\textbackslash t', `\textasciigrave\textasciigrave\textasciigrave' , `\ \ \ \ ' , `\#\#\#'\char125},\\
            & \texttt{\char123`\ \ \ \ \ \ \ \ ' , `\textbackslash t', `\textasciigrave\textasciigrave\textasciigrave' , `\ \ \ \ ' , `\#\#\#'\char125},\\
            & \texttt{\char123`\ \ \ \ \ \ \ \ ' , `\textbackslash t', `\textasciigrave\textasciigrave\textasciigrave' , `\ \ \ \ ' , `\#\#\#'\char125}\\
            \hline
            \multirow{3}{*}{$R_2$} & \texttt{\char123`\textasciigrave\textasciigrave\textasciigrave', '\textbackslash t', `\#\#\#', `\#\#', `\ \ \ \ \ \ \ \ '\char125},\\ & \texttt{\char123`\textasciigrave\textasciigrave\textasciigrave', '\textbackslash t', `\#\#\#', `\#\#', `\ \ \ \ \ \ \ \ '\char125},\\ & \texttt{\char123`\textasciigrave\textasciigrave\textasciigrave', '\textbackslash t', `\#\#\#', `\#\#', `\ \ \ \ \ \ \ \ '\char125}\\
            \hline
            \multirow{3}{*}{$R_3$} & \texttt{\char123`\textasciigrave\textasciigrave\textasciigrave', `\#\#', `\#\#\#', `\textbackslash t', `\#\#\#\#' \char125},\\ 
            & \texttt{\char123`\textasciigrave\textasciigrave\textasciigrave', `\#\#', `\#\#\#', `\textbackslash t', `\#\#\#\#' \char125},\\
            & \texttt{\char123`\textasciigrave\textasciigrave\textasciigrave', `\#\#', `\#\#\#', `\textbackslash t', `\#\#\#\#' \char125}\\
            \hline
            \multirow{3}{*}{$R_4$} & \texttt{\{` heavier', ` colleges', ` coloured', `akis', `ash'\}}, \\
            & \texttt{\{` heavier', ` colleges', `akis', ` coloured', `ash'\}},\\
            & \texttt{\{` heavier', ` colleges', `akis', `ash', `ni'\}} \\
            \hline
        \end{tabularx}
        \caption{Top-5 decoded tokens via Logit Lens at recurrent step 1}
        \label{tab:logit-r1}
    \end{table}

    \begin{table}[H]  % or [ht]
        \centering
        \begin{tabularx}{\columnwidth}{|c|X|}
            \hline
            \textbf{Recurrent Block} & \textbf{Top-5 Logit Lens Decoded Tokens Examples} \\
            \hline
            \multirow{3}{*}{$R_1$} & \texttt{\{`3', ` gre', `5', `2', `1'\}},\\
            & \texttt{\{`TV', `3', `5', `A', ` tru'\}},\\
            & \texttt{\{`5', `3', `TV', `MV', `None'\}}\\
            \hline
            \multirow{3}{*}{$R_2$} & \texttt{\{`3', `5', `2', ` gre', `8'\}},\\
            & \texttt{\{`3', `TV', `5', `6', ` gre'\}},\\
            & \texttt{\{`5', `3', `TV', ` inc', `None'\}}\\
            \hline
            \multirow{3}{*}{$R_3$} & \texttt{\{`3', `2', `8', `5', `1'\}},\\
            & \texttt{\{`3', `TV', `6', `5', `2'\}},\\
            &\texttt{\{`5', `3', `6', `TV', ` unity'\}}\\
            \hline
            \multirow{3}{*}{$R_4$}& \texttt{\{` weekend', `TED',} `\begin{CJK}{UTF8}{gbsn}得\end{CJK}', \texttt{`uru', ` gre'\}}, \\
            & \texttt{\{`ines', ` hav', `fly', ` classrooms', ` tru'\}},\\
            & \texttt{\{` stal', `igne', ` hav', `off', `ines'\}}\\
            \hline
        \end{tabularx}
        \caption{Top-5 decoded tokens via Logit Lens at recurrent step 8}
        \label{tab:logit-r8}
    \end{table}

    \begin{table}[H]
        \centering
        \begin{tabularx}{\columnwidth}{|c|X|}
            \hline
            \textbf{Recurrent Block} & \textbf{Top-5 Logit Lens Decoded Tokens Examples} \\
            \hline
            \multirow{3}{*}{$R_1$} 
                & \texttt{\{`3', `2', `1', `None', `|'\}} \\
                & \texttt{\{`TV', `3', `chi', `5', ` spa'\}} \\
                & \texttt{\{`5', `3', `1', ` answer', `None'\}} \\
            \hline
            \multirow{3}{*}{$R_2$} 
                & \texttt{\{`3', `2', `5', `|', `1'\}} \\
                & \texttt{\{`3', `5', `TV', `chi', `6'\}} \\
                & \texttt{\{`5', `3', `6', ` unity', ` answer'\}} \\
            \hline
            \multirow{3}{*}{$R_3$} 
                & \texttt{\{`2', `3', `8', `1', `5'\}} \\
                & \texttt{\{`3', `5', `6', `chi', `2'\}} \\
                & \texttt{\{`5', `3', ` unity', `6', `1'\}} \\
            \hline
            \multirow{3}{*}{$R_4$} 
                & \texttt{\{` optics', ` decor', ` doctors', ` po', ` chores'\}} \\
                & \texttt{\{`chi', `ani', `Factor', ` hav', `lag'\}} \\
                & \texttt{\{` inc', ` unity', ` friendships', `igne', ` impulse'\}} \\
            \hline
        \end{tabularx}
        \caption{Top-5 decoded tokens via Logit Lens at recurrent step 16}
        \label{tab:logit-r16}
    \end{table}
\pagebreak
Coda Lens:
\begin{table}[H]  % or [ht]
        \centering
        \begin{tabularx}{\columnwidth}{|c|X|}
            \hline
            \textbf{Recurrent Block} & \textbf{Top-5 Logit Lens Decoded Tokens Examples} \\
            \hline
            \multirow{3}{*}{$R_1$} & \texttt{\char123` plasma', ` sounds', ` draft', ` functions', `\textasciigrave\textasciigrave\textasciigrave'\char125},\\
            & \texttt{\char123` plasma', ` sounds', ` draft', `\textasciigrave\textasciigrave\textasciigrave', ` functions'\char125},\\
            & \texttt{\char123` plasma', ` sounds', ` draft', `\textasciigrave\textasciigrave\textasciigrave', `Answer'\char125}\\
            \hline
            \multirow{3}{*}{$R_2$} & \texttt{\char123`Answer', `-', `**', ` plasma', `\textasciigrave\textasciigrave\textasciigrave'\char125},\\ & \texttt{\char123`Answer', `-', `**', `\textasciigrave\textasciigrave\textasciigrave', ` plasma'\char125},\\ & \texttt{\char123`Answer', `-', `**', `\textasciigrave\textasciigrave\textasciigrave', ` plasma'\char125}\\
            \hline
            \multirow{3}{*}{$R_3$} & \texttt{\char123`Answer', `\textasciigrave\textasciigrave\textasciigrave', `-', `**', `\#\#\#\#'\char125},\\ 
            & \texttt{\char123`\textasciigrave\textasciigrave\textasciigrave', `Answer', `-', `**', `\#\#\#\#' \char125},\\
            & \texttt{\char123`Answer', `-', `\textasciigrave\textasciigrave\textasciigrave', `**', `You' \char125}\\
            \hline
            \multirow{3}{*}{$R_4$} & \texttt{\{`-', `Question', `Your', `Answer', `The'\}}, \\
            & \texttt{\{`-', `Question', `Your', `Answer', `The'\}},\\
            & \texttt{\{`-', `Question', `Your', `Answer', `The'\}} \\
            \hline
        \end{tabularx}
        \caption{Top-5 decoded tokens via Coda Lens at recurrent step 1}
        \label{tab:coda-r1}
    \end{table}

    \begin{table}[H]  % or [ht]
        \centering
        \begin{tabularx}{\columnwidth}{|c|X|}
            \hline
            \textbf{Recurrent Block} & \textbf{Top-5 Logit Lens Decoded Tokens Examples} \\
            \hline
            \multirow{5}{*}{$R_1$} & \texttt{\{` answer', ` answering', ` spa', ` Answers', ` answers'\}},\\
            & \texttt{\{` answer', ` greeting', ` spa', ` tru', ` product'\}},\\
            & \texttt{\{` answer', ` Answer', ` answering', ` highlighting', ` unity'\}}\\
            \hline
            \multirow{6}{*}{$R_2$} & \texttt{\{` answer', ` answering', ` answers', ` Answers', ` greeting'\}},\\
            & \texttt{\{` answer', ` answering', ` greeting', ` Answer', ` product'\}},\\
            & \texttt{\{` answer', ` answering', ` Answer', ` unity', ` answers'\}}\\
            \hline
            \multirow{3}{*}{$R_3$} & \texttt{\{`Answer', `\texttt{\textless|end\_turn|\textgreater}', ` answer', `3', `8'\}},\\
            & \texttt{\{'3', '6', 'Explanation', '\texttt{\textbackslash\textbackslash boxed}', '5'\}},\\
            &\texttt{\{'Explanation', '5', 'Answer', ' answer', '\texttt{\textbackslash\textbackslash boxed}'\}}\\
            \hline
            \multirow{3}{*}{$R_4$}& \texttt{\{`1', `4', `2', `8', `6'\}}, \\
            & \texttt{\{`6', `1', `3', `8', `7'\}},\\
            & \texttt{\{`5', `1', `6', `2', `7'\}}\\
            \hline
        \end{tabularx}
        \caption{Top-5 decoded tokens via Coda Lens at recurrent step 8}
        \label{tab:coda-r8}
    \end{table}

    \begin{table}[H]
        \centering
        \begin{tabularx}{\columnwidth}{|c|X|}
            \hline
            \textbf{Recurrent Block} & \textbf{Top-5 Logit Lens Decoded Tokens Examples} \\
            \hline
            \multirow{4}{*}{$R_1$} 
                & \texttt{\{` answering', ` answer', ` optics', ` spa', ` tweets'\}} \\
                & \texttt{\{` answer', ` spa', ` answers', ` greeting', ` alive'\}} \\
                & \texttt{\{` answer', ` answers', ` tru', ` clarification', ` spa'\}} \\
            \hline
            \multirow{5}{*}{$R_2$} 
                & \texttt{\{` answer', ` answering', ` Answer', ` radicals', ` answers'\}} \\
                & \texttt{\{` answer', ` answers', ` answering', ` greeting', ` Answer'\}} \\
                & \texttt{\{` answer', ` answers', ` Answer', ` answering', ` tru'\}} \\
            \hline
            \multirow{3}{*}{$R_3$} 
                & \texttt{\{`8', `Answer', `4', `9', `1'\}} \\
                & \texttt{\{`3', `6', `7', `4', `8'\}} \\
                & \texttt{\{`5', ` answer', `6', `7', `3'\}} \\
            \hline
            \multirow{3}{*}{$R_4$} 
                & \texttt{\{`1', `8', `-', `2', `6'\}} \\
                & \texttt{\{`6', `1', `8', `7', `3'\}} \\
                & \texttt{\{`6', `5', `1', `7', `2'\}} \\
            \hline
        \end{tabularx}
        \caption{Top-5 decoded tokens via Coda Lens at recurrent step 16}
        \label{tab:coda-r16}
    \end{table}

As seen in \cref{tab:logit-r8,tab:logit-r16,tab:coda-r8,tab:coda-r16}, As shown in \cref{tab:logit-r8,tab:logit-r16,tab:coda-r8,tab:coda-r16}, the proportion of numeric tokens shifts markedly around $R_4$, but in opposite directions for the two probing methods: it decreases under the Logit Lens and increases under the Coda Lens. This divergence reinforces our analysis and findings in \cref{sec:decode}.
\end{document}